\documentclass[letterpaper, 10 pt, conference]{ieeeconf} 

\IEEEoverridecommandlockouts                              
\usepackage{graphics} 
\usepackage{epsfig} 
\usepackage{mathptmx} 
\usepackage{times} 
\usepackage{amsmath} 
\usepackage{amssymb}  

\usepackage{algorithmic}
\usepackage[ruled,vlined,linesnumbered]{algorithm2e}
\usepackage{commath}

\newtheorem{definition}{Definition}

\usepackage{url}

\title{\LARGE \bf Can a Compact Neuronal Circuit Policy be Re-purposed to Learn Simple Robotic Control?}

\author{Ramin Hasani$^{1*}$, Mathias Lechner$^{2*}$, Alexander Amini$^{3}$, Daniela Rus$^{3}$ and Radu Grosu${^1}$
\thanks{*Equal Contributions}
\thanks{$^{1}$Technische Universit\"{a}t Wien (TU Wien), 1040 Vienna, Austria}%
\thanks{$^{2}$Institute of Science and Technology Austria (IST Austria)}%
\thanks{$^{3}$Massachusetts Institute of Technology (MIT), USA}%
}

\begin{document}

\maketitle
\thispagestyle{empty}
\pagestyle{empty}

\begin{abstract}

We propose a neural information processing system which is obtained by \textit{re-purposing} the function of a biological neural circuit model, to govern simulated and real-world control tasks. Inspired by the structure of the nervous system of the soil-worm, \emph{C. elegans}, we introduce \emph{Neuronal Circuit Policies} (NCPs), defined as the model of biological neural circuits reparameterized for the control of an alternative task. We learn instances of NCPs to control a series of robotic tasks, including the autonomous parking of a real-world rover robot. For reconfiguration of the \emph{purpose} of the neural circuit, we adopt a search-based optimization algorithm. Neuronal circuit policies perform on par and in some cases surpass the performance of contemporary deep learning models with the advantage leveraging significantly fewer learnable parameters and realizing interpretable dynamics at the cell-level.

\end{abstract}

\section{Introduction}
\begin{figure}[t!]
\centering
\includegraphics[width=0.50\textwidth]{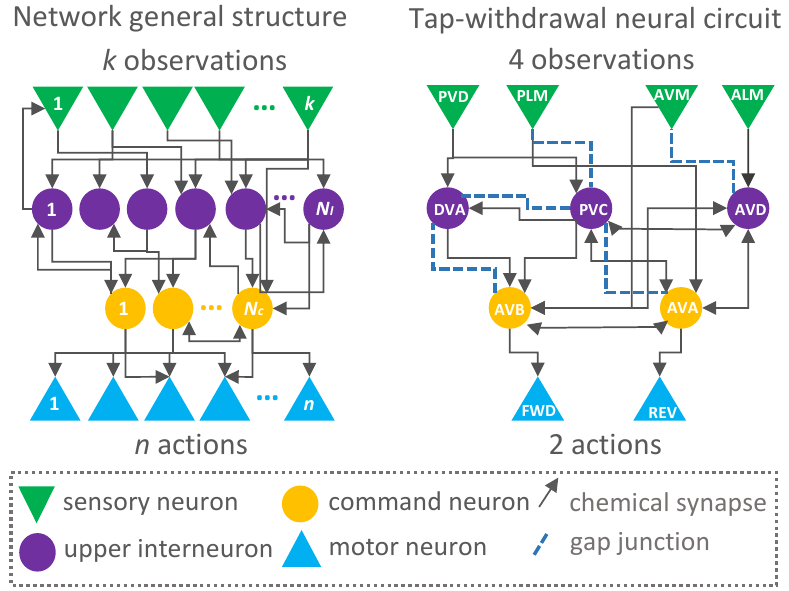}
 \caption{ Left: C. elegans' general neuronal circuit structure. Right: Tap-Withdrawal (TW) neural circuit schematic. Total number of interneurons $= N_i + N_C$. We preserve the TW circuit wiring topology, model its dynamics by computational models, and deploy a search-based reinforcement learning algorithm to control robots.}
 \label{figure1}
 \end{figure}

We wish to explore a new class of machine learning algorithms for robot control that is inspired by nature. Through natural evolution, the subnetworks within the nervous system of the nematode, \emph{C. elegans}, structured a near-optimal wiring diagram from the \textit{wiring economy principle}\footnote{Under proper functionality of a network, the wiring economy principle proposes that its morphology is organized such that the cost of wiring its elements is minimal \cite{perez2007optimally}.} perspective \cite{white86,perez2007optimally}. Its stereotypic brain composed of 302 neurons connected through approximately 8000 chemical and electrical synapses \cite{Chen2006}. \emph{C. elegans} exhibits distinct behavioral mechanisms to process complex chemical stimulations \cite{bargmann2006chemosensation}, avoid osmotic regions \cite{culotti1978osmotic}, sleep \cite{nichols2017global}, show adaptive behavior \cite{ardiel2010elegant}, perform mechanosensation \cite{chalfie1985neural}, and to control muscles \cite{wen2012proprioceptive}. The functions of many neural circuits within its brain have been identified \cite{Wicks95,Chalfie85,Li2012,nichols2017global}, and simulated \cite{m2017sim,hasani2017non,islam2016probabilistic,sarma2018openworm,gleeson2018c302}, which makes it a suitable model organism to be investigated, computationally. The general network architecture in \emph{C. elegans} establishes a hierarchical topology from sensory neurons through upper interneuron and command interneurons down to motor neurons (See Fig. \ref{figure1}A). In these neuronal circuits, typically interneurons establish highly recurrent wiring diagrams with each other while sensors and command neurons mostly realize feed-forward connections to their downstream neurons. An example of such a structure is a neural circuit shown in Fig. \ref{figure1}B, the Tap-withdrawal (TW) circuit \cite{rankin1990cad}, which is responsible for inducing a forward/backward locomotion reflex when the worm is mechanically exposed to touch stimulus on its body. The circuit has been characterized in terms of its neuronal dynamics \cite{Chalfie85}. The TW circuit comprises nine neuron classes which are wired together by means of chemical and electrical synapses. Behavior of the TW reflexive response is substantially similar to the control agent's reaction in some standard control settings such as a controller acting on driving an underpowered car, to go up on a steep hill, known as the \emph{Mountain Car} \cite{singh1996reinforcement}, or a controller acting on the navigation of a rover robot that plans to go from point A to B, on a planned trajectory. 

The biophysical neuronal and synaptic models express useful properties; 1) In addition to the nonlinearities exposed on the neurons' hidden state, synapses also are modeled by additional nonlinearity. This property results in realizing complex dynamics with a fewer number of neurons. 2) Their dynamics are set by grounded biophysical properties which ease the interpretation of the network hidden dynamics.

The trained networks obtained by learning the parameters of a fixed neural circuit structure are called \textit{neuronal circuit policies} (NCP). We experimentally investigate NCP's properties in terms, their learning performance, their ability to solve tasks in different RL domains, and ways to interpret their internal dynamics. For this purpose, we preserve the wiring structure of an example NCP (the TW circuit) and adopt a search-based optimization algorithm for learning the neuronal and synaptic parameters of the network. This paper contributes to the following:
\begin{itemize}
    \item Demonstration of a compact neuronal circuit policy (NCP) inspired by the brain of the \emph{C. elegans} worm, as an interpretable controller in various control settings.
    \item An optimization algorithm (Adaptive search-based) and experiments with NCPs in different control domains, including simulated and physical robots.
    \item Autonomous parking of a mobile robot by re-purposing a compact neuronal circuit policy.
    \item Interpretation of the internal dynamics of the learned policies. We introduce a novel computational method to understand continuous-time network dynamics. The technique (Definition 1) determines the relation between the kinetics of sensory/interneurons and a motor neuron's decision, We compute the magnitude of a neuron's contribution, (positive or negative), of these hidden nodes to the output dynamics in determinable phases of activity, during the simulation.
\end{itemize}

\section{Preliminaries}
In this section, we first briefly describe the structure and dynamics of the tap-withdrawal neural circuit. We then introduce the mathematical neuron and synapse models utilized to build up the circuit, as an instance of neuronal circuit policies.

\subsection{\textbf{Tap-Withdrawal Neural Circuit Revisit}}
A mechanically exposed stimulus (i.e. tap) to the petri dish in which the worm inhabits, results in the animal's reflexive response in the form of a forward or backward movement. This response has been named as the \emph{tap-withdrawal reflex}, and the circuit identified to underlay such behavior is known as the \emph{tap-withdrawal} (TW) neural circuit \cite{rankin1990cad}. The circuit is shown in Fig. \ref{figure1}B. It is composed of four sensory neurons, PVD and PLM (posterior touch sensors), AVM and ALM (anterior touch sensors), four interneuron classes (AVD, PVC, AVA and AVB), and two subgroup of motor neurons which are abstracted as a forward locomotory neurons, FWD, and backward locomotory neurons, REV. Interneurons recurrently synapse into each other with excitatory and inhibitory synaptic links. 

\subsection{\textbf{Neural circuit modeling revisit}}
Here, we briefly describe the neuron and synapse model model \cite{hasani2018liquid,lechner2019designing}, used to design neural circuit dynamics. Dynamics of neura
\begin{equation}
\label{eq:neuron_ode}
\begin{array}{lcl}
 \dot{V}_i(t) & = & [I_{i,L}\,{+}\,\sum_{j=1}^{n} \hat{I}_{i,j}(t) + \sum_{j=1}^{n} I_{i,j}(t)]\,{/}\,C_{i,m}\\[3mm]
 I_{i,L}(t) & = & \omega_{i,L}\,[E_{i,L}\,{-}\,V_i(t)]\\[3mm]
\hat{I}_{i,j}(t) & = & \hat{\omega}_{i,j}\,[V_j(t) - V_i(t)]\\[3mm]
I_{i,j}(t) & = & \omega_{i,j}\,[E_{i,j,R}\,{-}\,V_i(t)]\,g_{i,j}(t)\\[3mm]
g_{i,j}(t) & = & 1\,{/}\,[1+\mathrm{exp}(-\sigma_{i,j}\,(V_j(t)\,{-}\,\mu_{i,j}))]
\end{array}
\end{equation}
\noindent{}where $V_i(t)$ and $V_j(t)$ stand for the potential of the post and pre-synaptic neurons, respectively. $E_{i,L}$ and $E_{i,j}$ are the reversal potentials of the leakage and chemical channels. $I_{i,L}$, $\hat{I}_{i,j}$, and $I_{i,j}$ present the currents flowing through the leak channel, electric-synapse, and chemical-synapse, with conductances $\omega_{i,L}$, $\hat{\omega}_{i,j}$, and $\omega_{i,j}$, respectively. $g_{i,j}(t)$ is the dynamic conductance of the chemical-synapse, and $C_{i,m}$ is the membrane capacitance. $E_{i,j}$ determines the whether a synapse is inhibitory or excitatory.

For interacting with the environment, We introduced sensory and motor neuron models. A \textit{sensory component} consists of two neurons $S_p$, $S_n$ and an input variable, $x$. $S_p$ gets activated when $x$ has a positive value, whereas $S_n$ fires when $x$ is negative. The potential of the neurons $S_p$, and $S_n$, as a function of $x$, are defined by an affine function that maps the region $[x_{min}, x_{max}]$ of the system variable $x$, to a membrane potential range of $[-70mV, -20mV]$. (See the formula in Supplementary Materials Section 2).\footnote{Supplementary materials of the paper reside online at \url{https://github.com/mlech26l/neuronal_circuit_policies/tree/master/supplementary_material}} 
Similar to sensory neurons, a \textit{motor component} is composed of two neurons $M_n$, $M_p$ and a controllable motor variable $y$. Values of $y$ is computed by $y := y_p + y_n$ and an affine mapping links the neuron potentials $M_n$ and $M_p$, to the range $[y_{min}, y_{max}]$. (Formula in Supplementary Materials Section 2). FWD and REV motor classes (Output units) in Fig. \ref{figure1}B, are modeled in this fashion.

For simulating neural networks composed of such dynamic models, we adopted an implicit numerical solver \cite{Press2007}. Formally, we realized the ODE models in a hybrid fashion which combines both implicit and explicit Euler's discretization method \cite{lechner2019designing}. (See Supplementary Materials Section 3, for a concrete discussion on the model implementation, and the choice of parameters.) Note that the solver has to serve as a real-time control system, additionally. For reducing the complexity, therefore, our method realizes a fixed-timestep solver. The solver's complexity for each time step $\Delta_t$ is $\mathcal{O}(|\text{\# neurons}|+|\text{\# synapses}|)$. In the next section, we introduce the optimization algorithm used to reparametrize the tap-withdrawal circuit.

\section{Search-based Optimization Algorithm}
In this section we formulate a \emph{Reinforcement learning} (RL) setting for tuning the parameters of a given neural circuit to control robots. The behavior of a neural circuit can be expressed as a policy $\pi_\theta(o_i,s_i) \mapsto \langle a_{i+1}, s_{i+1}\rangle$, that maps an observation $o_i$, and an internal state $s_i$ of the circuit, to an action $a_{i+1}$, and a new internal state $s_{i+1}$. This policy acts upon a possible stochastic environment $Env(a_{i+1})$, that provides an observation $o_{i+1}$, and a reward, $r_{i+1}$. The stochastic return is given by $R(\theta) := \sum_{t=1}^{T} r_t$. The objective of the \emph{Reinforcement learning} is to find a $\theta$ that maximizes $\mathbb{E}\big(R(\theta)\big)$. 

Simple search based RL \cite{spall2005introduction}, as suggested in \cite{Salimans17}, in \cite{duan2016benchmarking}, and very recently in \cite{mania2018simple}, can scale and perform competitively with gradient-based approaches, and in some cases even surpass their performance, with clear advantages such as skipping gradient scaling issues. Accordingly, we adopted a simple search-based algorithm to train the neuronal policies.  

\begin{algorithm}[t]
\caption{Adaptive Random Search}
\label{alg1}
\begin{algorithmic}
\STATE {\bfseries Input:} A stochastic objective indicator $f$ and a starting parameter $\theta$, noise scale $\sigma$, adaption rate $\alpha\geq 1$
\STATE {\bfseries Output:} Optimized parameter $\theta$
\STATE $f_{\theta} \leftarrow f(\theta)$\;
\FOR{$k\leftarrow 1$ {\bfseries to} maximum iterations}
\STATE {$\theta' \leftarrow \theta + rand(\sigma)$;~~ $f_{\theta'} \leftarrow f(\theta')$;}
\STATE {\bfseries if}~~$f_{\theta'} < f_\theta$~~{\bfseries then}~~$\theta \leftarrow \theta'$;~~$f_\theta \leftarrow f_{\theta'}$;~~$i \leftarrow 0$;~~$\sigma \leftarrow \sigma \cdot \alpha$
{\bfseries else} $\sigma \leftarrow \sigma/\alpha$ {\bfseries end if} \;
\STATE $i\leftarrow i+1$\;
\STATE {\bfseries if}~~$i>N$~~{\bfseries then}~~$f_\theta \leftarrow f(\theta)$~~{\bfseries end if};
\ENDFOR
\STATE {\bfseries return} $\theta$\;
\end{algorithmic}
\end{algorithm}

Our approach combines a \emph{Adaptive Random Search} (ARS) optimization \cite{Rastrigin63}, with an \emph{Objective Estimate} (OE) function $f: \theta \mapsto \mathbb{R}^+$. The OE generates $N$ rollouts with $\pi_\theta$ on the environment and computes an estimate of $\mathbb{E}(R_{\theta})$ based on a filtering mechanism on these $N$ samples.
We compared two filtering strategies in this context; 1) taking the average of the $N$ samples, and 2) taking the average of the worst $k$ samples out of $N$ samples.

The first strategy is equivalent to the \emph{Sample Mean estimator} \cite{Salimans17}, whereas the second strategy aims to avoid getting misled by outlying high samples of $\mathbb{E}(R_{\theta})$. The objective for realizing the second strategy was the fact that a suitable parameter $\theta$ enforces the policy $\pi_\theta$ control the environment in a reasonable way even in challenging situations (i.e. rollouts with the lowest return). The algorithm is outlined in Algorithm \ref{alg1}. We treat the filtering strategy as a hyperparameter. 
\begin{table}[b!]
    \small
    \caption{Parameters and their bounds of a neural circuit}
    \centering
    \begin{tabular}{cccc}
        \textit{Parameter} & \textit{Value} & \textit{Lower bound} & \textit{Upper bound} \\\hline
        $C_{m}$ & Variable & 1mF & 1F \\
        $\omega_{L}$ & Variable & 50mS & 5S \\
        $E_{R}$ excitatory & 0mV & & \\
        $E_{R}$ inhibitory & -90mV & & \\
        $E_{L}$ & Variable & -90mV & 0mV \\
        $\mu$ & -40mV & &  \\
        $\sigma$ & Variable & 0.05 & 0.5  \\
        $\omega$ & Variable & 0S & 3S \\
        $\hat{\omega}$ & Variable & 0S & 3S \\
        \hline
    \end{tabular}
    \label{tab:parameter1}
\end{table}
\section{Experiments}
The goal of our experimentation is to answer the following questions: 
1) How would a neural circuit policy with a preserved biological connectome, perform in basic standard control settings, compared to that of a randomly wired circuit? 2) When possible, how would the performance of our learned circuit compare to the other methods? 3) Can we transfer a policy from a simulated environment to a real environment? 4) How can we interpret the behavior of the neural circuit policies? 

We use four benchmarks for measuring and calibrating the performance of this approach including one robot application to parking for the TW sensory/motor neurons and then deployed our RL algorithm to learn the parameters of the TW circuit and optimize the control objective. The environments include I) Inverted pendulum of Roboschool \cite{schulman2017}, II) Mountain car of OpenAI Gym, III) Half-CHeetah from Mujoco, and IV) Parking a real rover robot with a transferred policy from a simulated environment. The code is available online.
\footnote{Code for all experiments is available online at: \url{https://github.com/mlech26l/neuronal_circuit_policies}} 
The TW neural circuit (cf. Fig. \ref{figure1}B) allows us to incorporate four input observations and to take two output control actions. We evaluate our NCP in environments of different toolkits on a variety of dynamics, interactions, and reward settings. 

\subsection{\textbf{How to map NCPs to environments?}}
The TW neural circuit is shown in Fig. \ref{figure1}B, contains four sensory neurons. It, therefore, allows us to map the circuit to four input variables. Let us assume we have an inverted pendulum environment which provides four observation variables The position of the cart $x$, together with its velocity $\dot{x}$, the angle of the pendulum $\varphi$.\footnote{Remark: The environment further splits $\varphi$ into $sin(\varphi)$ and $cos(\varphi)$ to avoid the $2\pi \rightarrow 0$ discontinuity} along with its angular velocity $\dot{\varphi}$. Since the main objective of the controller is to balance the pendulum in an upward position and make the car stay within the horizontal borders, we can feed $\varphi$ (positive and negative values), and $x$ (positive and negative), as the inputs to the sensors of the TW circuit. Control commands can be obtained from the motor neuron classes, FWD and REV. Likewise, any other control problem can be feasibly mapped to an NCP. We set up the search-based RL algorithm to optimize neurons' and synapses' parameters $\omega, \hat{\omega}, \sigma, C_m, E_{L}$ and $G_{L}$, within their corresponding range, shown in Table \ref{tab:parameter1}. A video of different stages of the learned neuronal circuit policy for the inverted pendulum can be viewed at \url{https://youtu.be/cobEtJVw3A4}

\begin{figure}[t!]
\centering
\includegraphics[width=0.5\textwidth]{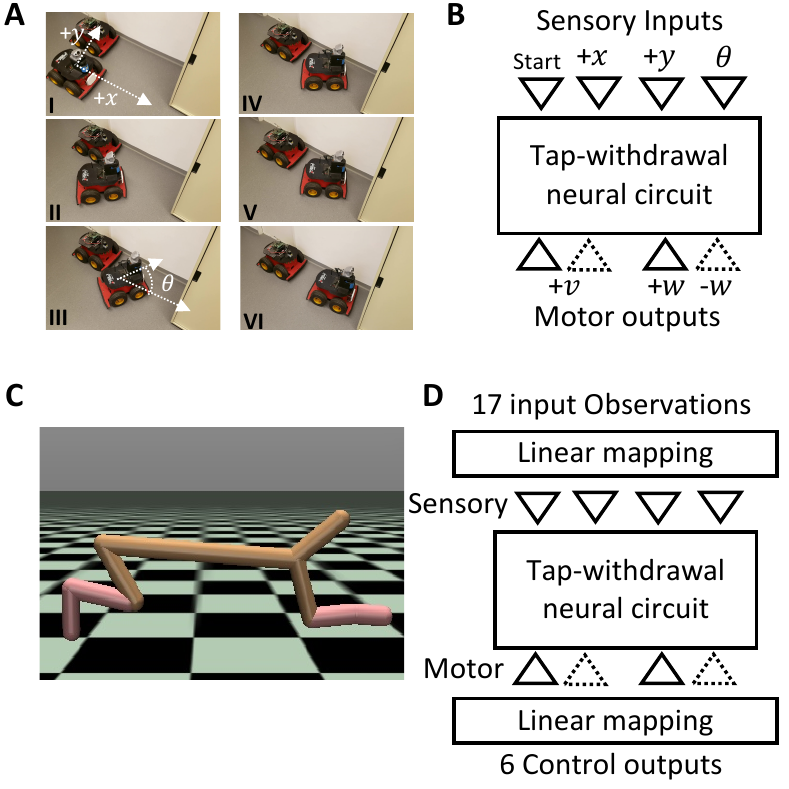}
 \caption{ Mapping the environments to the TW circuit in A) Parking task, B) mapping for the parking. C) half-cheetah, and C) mapping for the half-cheetah experiment. See Table S3 in the Supplementary Material for more details.}
 \label{experiments}
 \end{figure}
In a simulated Mountaincar experiment, the environmental variables are the car's horizontal position, $x$, together with its linear velocity. The control signal applies force to the car to build up momentum until finally reaching the top of the hill. The TW circuit can then be learned by the search-based RL algorithm. A video illustrating the control of the car at various episodes during the optimization process can be viewed at \url{https://youtu.be/J7vXFsZz7EM}.

\subsection{\textbf{Scale the functionality of NCPs to environments with larger observation spaces}} We extend the application of the TW circuit as an instance of neuronal circuit policies, to handle tasks with more observation variables. We choose the HalfCheetah-v2 test-bed of Mujoco. The environment consists of 17 input and six output variables. We add a linear layer that maps an arbitrary number of input variables to two continuous variables that are then fed into the four sensory neurons of the TW circuit, as shown in Fig. \ref{experiments}D. Similarly, we add a linear layer that maps the neuron potentials of the two motor neurons to the control outputs. A video of this experiment can be viewed at \url{https://youtu.be/3EhHLINROo4}.

\subsection{\textbf{Transfer learned NCPs to control real robot}}
In this experiment, we generalized our TW neuronal circuit policy to a real-world control test-bed. We let the TW circuit learn to park a rover robot on a determined spot, given a set of checkpoints form a trajectory, in a deterministic simulated environment. We then deploy the learned policy on a mobile robot in a real environment shown in Fig. \ref{experiments}A. The key objective here is to show the capability of the method to perform well in a transformation from a simulated environment to a real setting. For doing this, we developed a \emph{custom deterministic simulated RL environment}.

The rover robot provides four observational variables (start signal, position ($x$, $y$) and angular orientation $\theta$), together with two motor actions (linear and angular velocity, $v$ and $w$). We mapped all four observatory variables, as illustrated in Fig. \ref{experiments}B, to the sensors of the TW. Note that here the geometric reference of the surrounding space is set at the initial position of the robot. Therefore, observation variables are positive. We mapped the linear velocity (which is a positive variable throughout the parking task) to one motor neuron and the same variable to another motor neuron. We determined two motor neurons for the positive and negative angular velocity. (See Table S3 in Supplementary for mapping details).\footnote{Supplementary materials of the paper reside online at \url{https://github.com/mlech26l/neuronal_circuit_policies/tree/master/supplementary_material}} This configuration implies that the command neuron, AVA, controls two motor neurons responsible for the turn-right and forward motion-primitives, and AVB to control the turn-left and also forward motor neurons.

\textbf{Optimization setup for the parking task --}
A set of checkpoints on a pre-defined parking trajectory were determined in the custom simulated environment. For every checkpoint, a deadline was assigned. At each deadline, a reward was given as the negative distance of the rover to the current checkpoint. The checkpoints are placed to resemble a real parking trajectory composed of a sequence of motion primitives: Forward, turn left, forward, turn right, forward, and stop. We then learned the TW circuit by the RL algorithm. The learned policy has been mounted on a Pioneer AT-3 mobile robot and performed a reasonable parking performance. The Video of the performance of the TW neuronal circuit policy on the parking task can be viewed at \url{https://youtu.be/p0GqKf0V0Ew}.

\section{Experimental Evaluation}
In this section, we thoroughly assess the results of our experimentation. We qualitatively and quantitatively explain the performance of our neuronal circuit policies. We then benchmark our results where possible, with the existing methods and describe the main attributes of our methodology. Finally, we quantitatively interpret the dynamics of the learned policies.

\subsection{\textbf{Are NCPs better than random circuits?}}
We performed an experiment in which we designed circuits with randomly wired connectomes, with the same number of neurons and synapses used in the TW circuit. (11 neurons, 28 synapses). Connections between each neuron pair are set randomly by a random choice of the synapse type (excitatory, inhibitory or gap junction) with a simple rule that no synapse can be fed into
a sensory neuron, which is a property of NCPs as well. The random circuits were then trained over a series of control tasks described earlier, and their performance is reported in Table \ref{random}. We observe that NCPs significantly outperform the randomly wired networks, which is empirical evidence for the applicability of the NCPs.

\begin{table}[h!]
\small
\centering
\caption{NCP versus random circuits - Results are computed for 10 random circuits, and 10 weight initialization runs of the TW circuit - High standard deviations are due to the inclusion of unsuccessful attempts of each type of network.}
\begin{tabular}{c|cc}
\hline
\textbf{Env / Method} & Random Circuit & \textbf{NCP} \\
\hline
Inverted Pendulum & 138.1$\pm$ 263.2 & \textbf{866.4 $\pm$418}\\
Mountain car & 54 $\pm$44.6 & \textbf{91.5$\pm$6.6} \\
Half-Cheetah & 1742.9$\pm$642.3 & \textbf{2891.4$\pm$1016}\\
\hline 
\end{tabular}
\label{random}
\end{table}

\subsection{Performance} 
\begin{figure}[t!]
\centering
\includegraphics[width=0.49\textwidth]{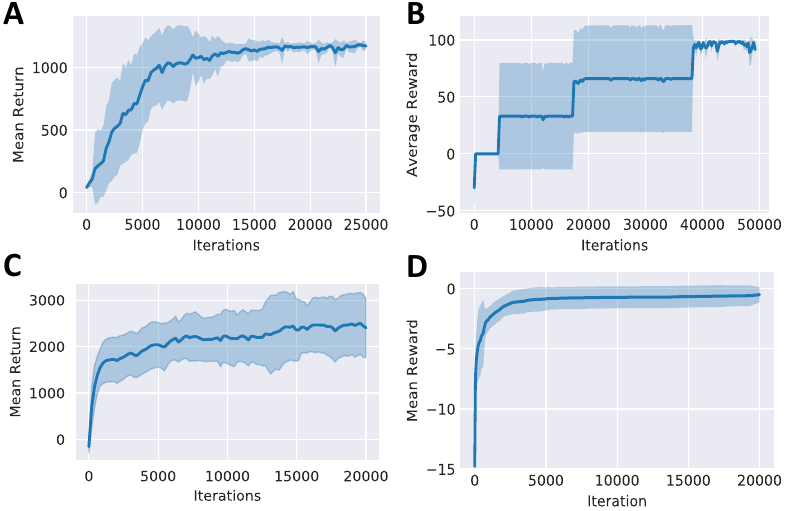}
 \caption{Learning curves for the TW circuit in standard RL tasks. A) Inverted pendulum B) Mountain car (OpenAI Gym) C) Half-Cheetah D) The parking task. The shadows around the learning curves represent the standard deviation in learning each task for 10-times repetitions.}
 \label{learning_curves}
 \end{figure}

The training algorithm was able to solve all the tasks, after a reasonable number of iterations, as shown in the learning curves in Fig. \ref{learning_curves}A-D.
Jumps in the learning curves of the mountain car (Fig. \ref{learning_curves}B) are the consequence of the sparse reward. For the deterministic parking trajectory, the learning curve converges in less than 5000 iterations. 

\subsection{\textbf{How does NCP + random search compares with policy gradient based RL algorithms?}}
NCPs + Random search algorithm demonstrates comparable performance to the state-of-the-art policy gradient RL algorithms such as Proximal Policy optimization (PPO) \cite{schulman2017}, and advantage actor critic (A2C) \cite{mnih2016asynchronous}. Table \ref{comparision} reports the performance of the mentioned algorithms compared to NPC+RS.

\begin{table}[h!]
\small
\centering
\caption{Comparison of NCP to artificial neural networks with policy gradient algorithms}
\begin{tabular}{c|cc}
\hline
\textbf{Method} & {Inverted Pendulum} & {MountainCar}\\
\hline
MLP + PPO \cite{schulman2017}& 1187.4$\pm$51.7 & 94.6$\pm$1.3\\
MLP + A2C \cite{mnih2016asynchronous}& 1191.2$\pm$45.2 & 86.4$\pm$18.3 \\
NCP + RS (ours) & 1168.5$\pm$21.7 & 91.5$\pm$6.6\\
\hline 
\end{tabular}
\label{comparision}
\end{table}

\subsection{\textbf{How does NCP compare to deep learning models?}}
The final return values for the basic standard RL tasks (provided in Table \ref{training_results}), matches that of conventional policies \cite{heidrich2008variable}, and the state-of-the-art deep neural network policies learned by many RL algorithms \cite{schulman2017,NIPS2017_6692}. We compared the performance of the learned TW circuit to long short-term memory (LSTM) recurrent neural networks \cite{hochreiter1997long}, multi-layer peceptrons (MLP), and random circuits. We select the same number of cells (neurons) for the LSTM and MLP networks, equal to size of the tap-withdrawal circuit. LSTM and MLP networks are fully connected while the TW circuit realizes a $77 \%$ network sparsity. In simple tasks experiments the TW circuit performs in par with the MLP and LSTM networks, while in the HalfCheetah experience, it significantly achieves a better performance. Results are summarized in Table \ref{training_results}.
\begin{table*}[h]
\centering
\caption{Compare NCP with deep learning models. numbers show the Mean, standard deviation, and success rate for 10 runs. $N=10$}
\resizebox{1.\linewidth}{!}{
\begin{tabular}{c|ccc|c}
\hline
\textbf{Agent} & \textbf{Inverted Pendulum} & \textbf{Mountaincar} & \textbf{HalfCheetah} & Sparsity \\
\hline
LSTM & 629.01 $\pm$ 453.1 (40.0\%) & \textbf{97.5} $\pm$ 1.25 (100.0\%) & 1588.9 $\pm$ 353.8 (10.0\%) & 0\% (fully connected)\\
MLP & \textbf{1177.49} $\pm$ 31.8 (100.0\%) & 95.9 $\pm$ 1.86 (100.0\%) & 1271.8 $\pm$ 634.4 (0.0\%)& 0\% (fully connected) \\
NCP (ours) & \textbf{1168.5}$ \pm$ 21.7 (90.0\%) & 91.5 $\pm$ 6.6 (80.0\%) & \textbf{2587.4} $\pm$ 846.8 (72.7\%) & \textbf{77\% (28 synapses)}\\
Random circuit & 138.10 $\pm$ 263.2 (10.00\%) & 54.01$\pm$ 44.63 (50.0\%) & 1743.0 $\pm$ 642.3 (50.0\%) & 77\% (28 synapses)\\\hline
\end{tabular}
}
\label{training_results}
\end{table*}
\subsection{\textbf{Interpretability of the neuronal circuit policies}}
\begin{figure*}[h!]
\centering
\includegraphics[width=1\textwidth]{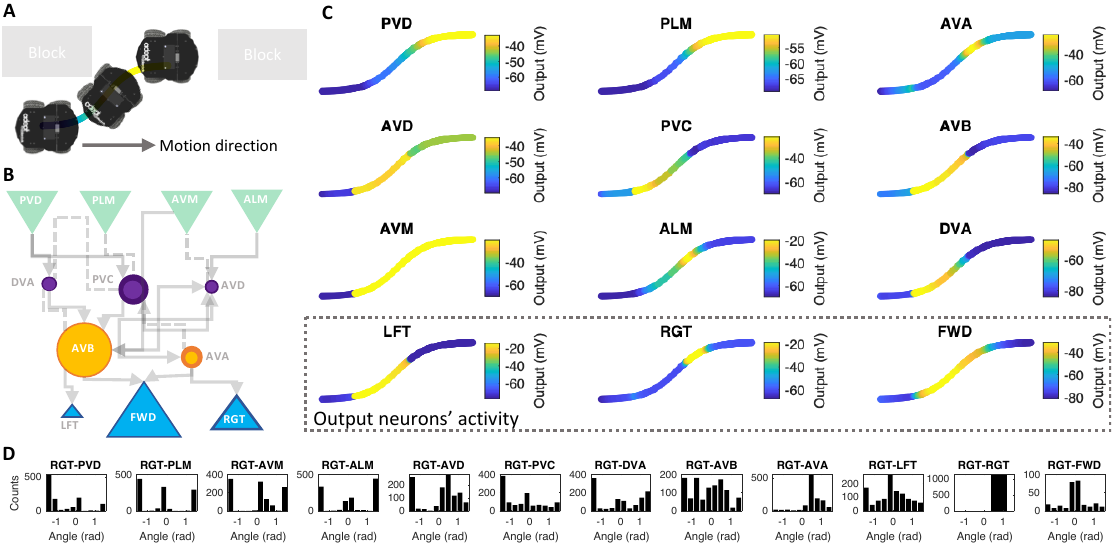}
 \caption{Interpretability analysis of the parking task. A) The parking trajectory. B) TW circuit drawn with the range of possible variations of the individual neuron's time-constants; the radius of the darker color circle for each neuron corresponds to the range within which the time-constant varies between $\tau_{min}$ and $\tau_{max}$ while the robot performs the parking. (Values in Supplementary Materials, Section 8.). C) Projection of individual neuron's output over the parking trajectory. The plots demonstrate when neurons get activated while the rover is performing the parking task. D) Histogram of the slopes in manifolds' point-pair angles for a motor neuron, in the parking task. (See Supplementary Materials Section 7, for full circuit's analyses, in other experiments.)}
 \label{tau}
 \end{figure*}

In this section, we introduce a systematic method for interpreting the internal dynamics of an NCP.  The technique determines how the kinetics of sensory neurons and interneurons relate to a motor neuron's decision.
Fig. \ref{tau}B illustrates how various adaptive time-constants are realized in the parking environment. Interneurons (particularly PVC and AVA) change their time-constants significantly compared to the other nodes. This corresponds to their contribution to various dynamical modes and their ability to toggle between dynamic phases of an output decision. Fig. \ref{tau}C visualizes the activity of individual TW  neurons (lighter colors correspond to a more activation phase) over the parking trajectory. It becomes qualitatively explainable how individual neurons learned to contribute to performing autonomous parking. For instance, AVA which is the command neuron for turning the robot to the right-hand-side (Motor neuron RGT) while it is moving, gets highly activated during a right turn. Similarly, AVB and LFT neurons are excited during a left-turning phase. (See Fig. \ref{tau}C).
Next, we formalize a quantitative measure of a NCP element's contribution to its output decision. 

\begin{definition}
    Let $I =[0,T]$ be a finite simulation time of a neuronal circuit policy with $k$ input neurons, $N$ interneurons and $n$ motor neurons, (Shown in Fig. \ref{figure1}A), acting in an RL environment. For every neuron-pair ($N_i$,$n_j$), ($N_i$, $N_j$) and ($k_i$, $n_j$) of the circuit, In a cross-correlation space, let $S = \{s_1, .... s_{T-1}\}$ be the set of the gradients amongst every consecutive pair points, and $\Omega = \{\arctan(s_1), ..., \arctan(s_{T-1}\}$ be the set of all corresponding geometrical angles, bound in a range $[-\frac{\pi}{2}, \frac{\pi}{2}]$. Given the input dynamics, we quantify the way sensory neurons and interneurons contribute to motor neurons' dynamics, by computing the histogram of all $\Omega$s, with a bin-size equal to $l$ (i.e. Fig \ref{tau}D), as follows:
    \begin{itemize}
        \vspace{-1mm}\item If sum of bin-counts of all $\Omega > 0$, is more than half of the sum of bin-counts in the $\Omega < 0$, the overall contribution of $N_i$ to $n_j$ is considered to be positive.\vspace{-1mm}
        \item If sum of bin-counts of all $\Omega < 0$, is more than half of the sum of bin-counts in the $\Omega > 0$, the overall contribution of $N_i$ to $n_j$ is negative,\vspace{-1mm}
        \item Otherwise, $N_i$ contributes in phases (switching between antagonistic and phase-alighted) activity of $n_j$, on determinable distinct periods in $I$.\vspace{-1mm}
    \end{itemize}
    \label{def1}
\end{definition}

To exemplify the use of the proposed interpretability method, let us consider the neuronal activity of a learned circuit driving a rover robot autonomously on a parking trajectory. 
Fig. \ref{tau}D presents the histograms computed by using Definition 1,  for the RGT motor neuron dynamics (ie. the neuron responsible for turning the robot to the right) in respect to that of other neurons. Based on Definition 1, we mark AVM, AVD, AVA as positive contributors to the dynamics of the RGT motor neuron. We determine PVD, PLM, and PVC as antagonistic contributors. Neurons such as DVA and AVB realized phase-changing dynamics where their activity toggles between a positive and negative correlations, periodically. (For the analysis of the full networks' activities visit Supplementary Materials Section 7).\footnote{Supplementary materials of the paper reside online at \url{https://github.com/mlech26l/neuronal_circuit_policies/tree/master/supplementary_material}} 

Such analysis is generalizable to the other environments too. (See Supplementary Materials Section 7). In that case, the algorithm determines principal neurons in terms of neuron's contribution to a network's output decision in computable intervals within a finite simulation time. 

\section{Conclusions}
We showed the performance of neuronal circuit policies in control environments. We empirically demonstrated that the sub-networks taken directly from the nervous system of the small species perform significantly better than randomly wired circuits. We selected the tap-withdrawal neural circuit of the nematode worm as an example NCP, repurposed its parameters by a search based RL algorithm and governed simulated control tasks and generalized to a real-life robotic application. We experimentally demonstrated the interpretable control performance of the learned circuits in action and introduced a quantitative method to explain networks' dynamics. The proposed method can also be utilized as a building block for the interpretability of recurrent neural networks, which despite a couple of fundamental studies \cite{karpathy2015visualizing,chen2016infogan,olah2018building,hasani2019response}, is still a grand challenge to be addressed. 
Finally, we open-sourced our methodologies, to encourage other researchers to explore the attributes of neuronal circuit policies further and apply them to other control and reinforcement learning domains. 

\section*{ACKNOWLEDGMENT}
R.H., and R.G. are partially supported by the Horizon-2020 ECSEL Project grant No. 783163 (iDev40), and the Austrian Research Promotion Agency (FFG), Project No. 860424. T.H. was supported in part by the Austrian Science Fund (FWF) under grant Z211-N23 (Wittgenstein Award). A.A. is supported by the National Science Foundation Graduate Research Fellowship under Grant No. 1122374. R.H., and D.R. are partially supported by the Boeing Company. 


\end{document}